%% file: main.tex
\documentclass[]{spie}  %>>> use for US letter paper

\usepackage{amsmath,amsfonts,amssymb}
\usepackage{subcaption}
\usepackage{graphicx}
\usepackage{cite} 
\usepackage{epsfig}
\usepackage{nccmath}
\usepackage{xcolor,colortbl}
\usepackage{tabularx}
\usepackage{relsize}
\usepackage{pifont}

\usepackage{booktabs} 
\usepackage{multirow}
\usepackage{multicol}
\usepackage{adjustbox}
\usepackage{float}
\usepackage{makecell}
\usepackage{tabu}
\usepackage[colorlinks=true, allcolors=blue]{hyperref}
\usepackage[capitalize]{cleveref}
\usepackage{algorithm}

\title{3D Banff Glomerulitis Scoring for Kidney Transplant Pathology: Feasibility Study for Volumetric Quantification}

\author[a]{Yanfan Zhu}
\author[a]{Junchao Zhu}
\author[b]{Ruining Deng}
\author[c]{Shilin Zhao}
\author[c]{Yu Wang}
\author[d]{Yaohong Wang}
\author[a]{Yuechen Yang}
\author[a]{Chongyu Qu}
\author[a]{Zhengyi Lu}
\author[a]{Juming Xiong}
\author[c]{Mengmeng Yin}
\author[c]{Yihan Wang}
\author[c]{Andrew J. Rauch}
\author[c]{Haichun Yang}
\author[a]{Yuankai Huo}

\affil[a]{Vanderbilt University, Nashville TN 37235, USA}
\affil[b]{Weill Cornell Medicine, New York, NY 10021, USA}
\affil[c]{Vanderbilt University Medical Center, Nashville TN 37232, USA}
\affil[d]{UT MD Anderson Cancer Center,TX 77030, USA}
\authorinfo{Further author information: (Send correspondence to Yuankai Huo)\\
E-mail: yuankai.huo@vanderbilt.edu}

\begin{document}
\maketitle

\begin{abstract}
Quantifying glomerular inflammation in kidney-transplant biopsies is traditionally performed on single whole-slide sections using the Banff "most-severe section" rule, despite the inherently three-dimensional nature of lesion distribution. This 2D based assessment might lead to instability when inflammation varies across slices. This paper reports a feasibility study examining the technical feasibility and clinical relevance of extending Banff scoring into three dimensions. To this end, we propose a 3D Banff lesion-scoring framework that aligns serial sections, reconstructs glomerular identities across depth, and integrates inflammatory-cell counts in 3D under three scoring strategies: Compatible, Volumetric, and Deduplicated Volumetric. In the experiments, glomerulitis (\textit{g}-scores) was used as an example Banff metric and applied to multi-section renal allograft biopsies. Our findings indicate that 3D Banff \textit{g}-scores are more consistent across sections and, under the semi-automatic setting, the Volumetric strategy correlates more strongly with clinical biomarkers than traditional 2D scores. These results show that 3D volumetric quantification offers promising added value, underscoring the potential benefit of 3D-aware Banff scoring for kidney transplant pathology.

\end{abstract}


\input{sections/1_introduction}

\input{sections/2_methods}
\input{sections/3_experiments}
\input{sections/4_results}
\input{sections/5_conclusion}

\acknowledgements
% equivalent to \section*{ACKNOWLEDGMENTS}       
\begin{flushleft}
This research was supported by NIH R01DK135597 (Huo), DoD HT9425-23-1-0003 (HCY), NSF 2434229 (Huo) and KPMP Glue Grant. This work was also supported by Vanderbilt Seed Success Grant and Vanderbilt-Liverpool Seed Grant. This research was also supported by NIH grants R01EB033385, R01DK132338. We extend gratitude to NVIDIA for their support by means of the NVIDIA hardware grant. 

\end{flushleft}
\bibliographystyle{spiebib} 
\bibliography{references}

\end{document}

%% file: sections/1_introduction.tex
\section{Introduction}

\begin{figure*}[t]
    \centering
    \includegraphics[width=\textwidth]{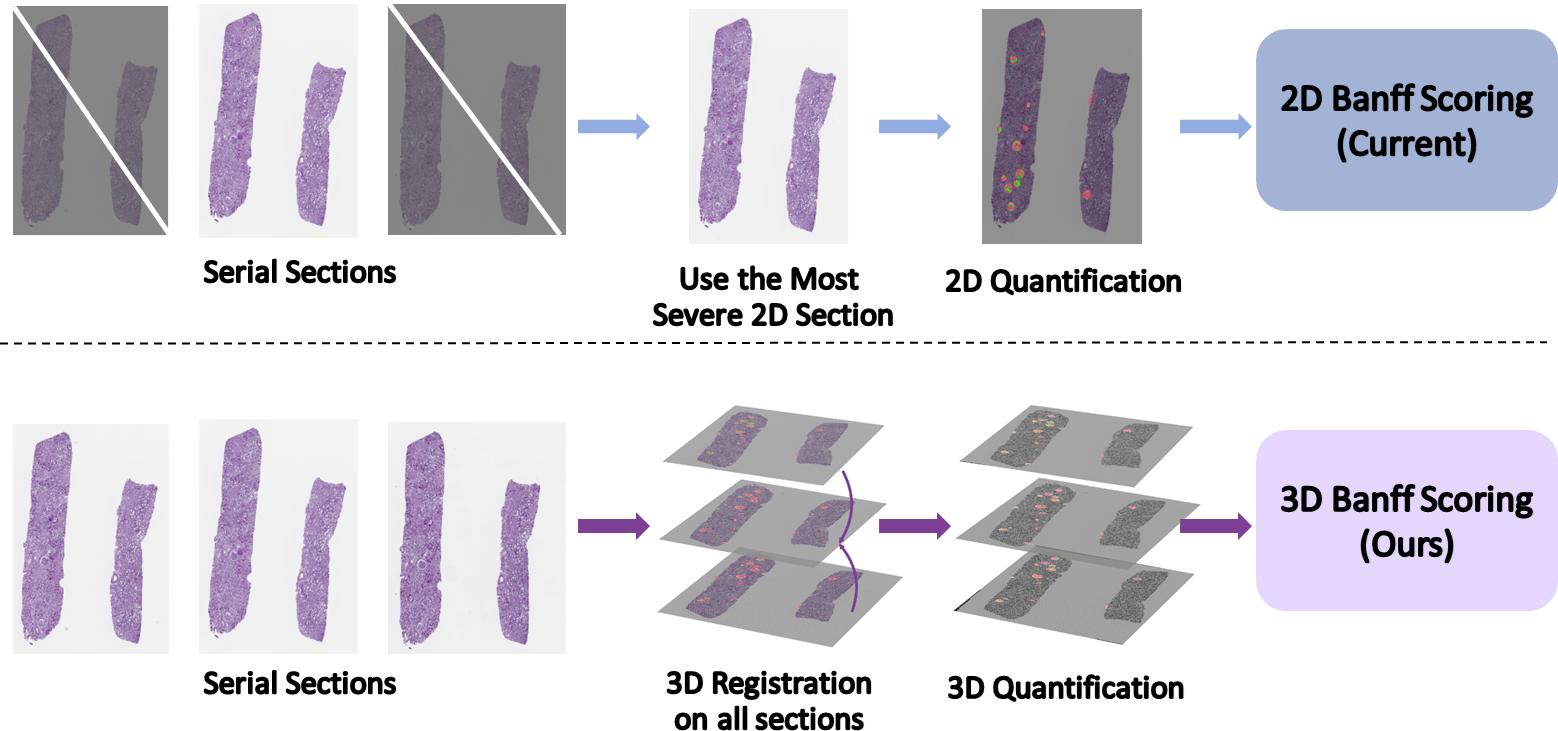}
    \caption{\textbf{2D Banff scoring vs.\ 3D Banff scoring.} \textbf{Upper panel:} the 2D Banff pipeline analyzes each section independently and assigns the final \textit{g}-score based on the most severe single section. This ``worst-section'' rule is sensitive to sampling variability and ignores cross-section correspondence. \textbf{Lower panel:} the proposed 3D Banff framework aligns serial sections and integrates information across depth for volumetric \textit{g}-score computation.}
    \label{fig:overview}
\end{figure*}

Renal allograft biopsy is fundamental to diagnosing rejection and determining post-transplant management. The Banff Classification of Allograft Pathology provides the international standard for evaluating renal transplant biopsies, using a set of semiquantitative lesion scores—including interstitial inflammation (\textit{i}), tubulitis (\textit{t}), peritubular capillaritis (\textit{ptc}), intimal arteritis (\textit{v}), and glomerulitis (\textit{g})—to diagnose and grade rejection \cite{Farris2025_AI_KidneyBiopsyReview,banffrepo2023,zhu2025closewelimitationsprogress}.
Although a single biopsy block is routinely cut into 20--30 serial sections, each representing a slightly different tissue depth, current clinical practice assigns the \textit{g}-score using only one two-dimensional (2D) histology section—the slice showing the most severe inflammation. This worst-section approach aims to avoid underestimating rejection but is inherently sensitive to sampling variation, sectioning depth, and local tissue distortion. Clinically, glomerular inflammation is a three-dimensional (3D) biological phenomenon, yet its scoring relies on a single 2D snapshot. This mismatch between disease geometry and the evaluation protocol leads to variability in diagnosis and reduces interpretability for both pathologists and clinicians~\cite{Loupy2022_Banff30yrs,Huang2023_BanffReproducibility}.

Recent advances in computational pathology have produced accurate models for segmenting glomeruli\cite{deng2023omnisegscaleawaredynamicnetwork,deng2025irsincrementalrelationshipguidedsegmentation,Jiang2021_GlomeruliInstanceSeg,yu2024gloinonev2holisticidentificationglomerular} and detecting inflammatory cells\cite{Deotale2025_MonkeyKidneyInflammation,Mengmonkeychallenge2023} on whole slide images (WSIs)\cite{LABRIFFE20222821,Yi2024_DL_ProcurementBiopsy}. However, nearly all existing systems operate strictly in 2D: each WSI is processed independently; predictions are not aligned or reconciled across serial sections; and case-level scoring still selects the single most severe slice, replicating the clinical worst-section paradigm~\cite{hermsen2019deep,Kers2022_DLKidneyTransplant,Hermsen2022_CNNKidneyTransplant}. These approaches fail to reconstruct the 3D topology of injury, do not assign consistent glomerulus identity across depth, and cannot quantify volumetric inflammatory distribution. From a clinical standpoint, this limits robustness, explainability, and trustworthiness, all essential for adoption of AI tools in routine transplant workflows.

We propose a shift from 2D slice-based evaluation to biopsy-level 3D volumetric scoring. Our framework aligns serial histology sections, reconstructs glomeruli in 3D, aggregates inflammatory-cell detections, and produces a biopsy-level Banff \textit{g}-score derived from the full depth of the tissue block. This requires addressing several technical challenges: nonlinear distortions and missing sections that hinder registration; substantial variation in staining and section quality; identity tracking of glomeruli across levels; and designing an interpretable and clinically grounded method for compatible and volumetric aggregation. To overcome these challenges, we developed a unified computational pipeline that integrates robust 2D feature extraction, serial-section registration, glomerulus identity tracking, and 3D \textit{g}-score computation (Figure~\ref{fig:overview}). Our contributions are as follows:

\begin{itemize}
  \item This paper presents a feasibility study examining the technical feasibility and clinical relevance of extending Banff scoring from 2D to 3D.
  \item We propose a computational framework for the quantitative measurement of the widely used Banff \textit{g}-score, further extending it to 3D through the integration of AI-based section detection, glomerular segmentation, inflammatory-cell detection, and structural alignment across serial histological sections.
  \item We formulate three biopsy-level \emph{3D Banff \textit{g}-scoring} strategies---Compatible, Volumetric, and Deduplicated Volumetric---that capture different aspects of the 3D inflammatory distribution, and benchmark their performance against 2D Banff scoring.
\end{itemize}

\begin{figure*}[t]
    \centering
    \includegraphics[width=0.95\textwidth]{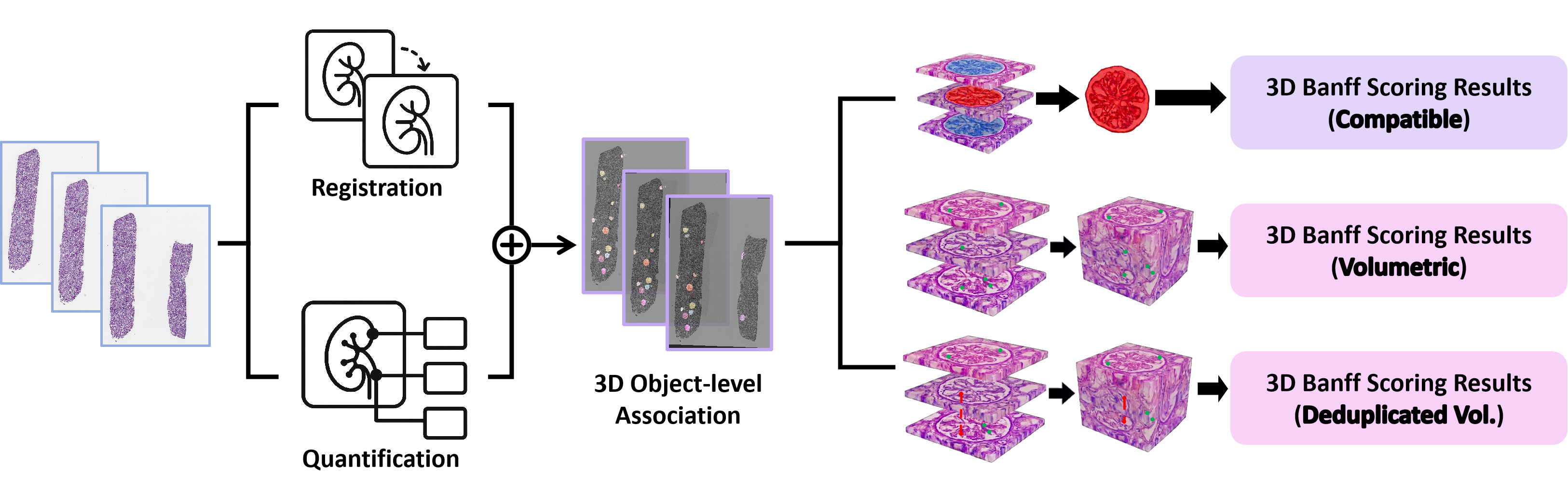}
    \caption{\textbf{Proposed 3D Banff scoring pipeline.}
    Serial sections are registered and quantified in parallel. Their outputs are combined to associate glomeruli across depth, followed by Compatible, Volumetric, and Deduplicated Volumetric scoring.}
    \label{fig:pipeline}
\end{figure*}

%% file: sections/2_methods.tex
\section{Methods}
\label{sec:methods}
\subsection{Pipeline Overview and 2D Banff Scoring}

Figure~\ref{fig:pipeline} illustrates the complete workflow of the proposed 3D Banff lesion scoring pipeline. Raw serial-section WSIs first enter two parallel modules: \textbf{Registration}, which aligns all sections from a biopsy into a common spatial frame, and \textbf{Quantification}, which uses deep-learning models to extract per-section glomerular and inflammatory-cell features. The outputs of these modules are then combined to establish consistent 3D object associations for depth-wise tracking, whereby corresponding glomeruli across sections are linked based on spatial overlap, geometric continuity, and instance-level matching. These associations produce coherent 3D glomerular entities with aggregated inflammatory-cell detections along depth, from which three 3D Banff scoring strategies are derived: \textbf{Compatible} (Banff worst-section rule in 3D), \textbf{Volumetric} (depth-normalized cumulative cell count), and \textbf{Deduplicated Volumetric} (unique-cell volumetric count after cross-section deduplication).

As a computational baseline consistent with conventional pathology workflows, we implement an automated per-section \textit{g}-score following the classical Banff definition of glomerulitis. Let $\{R_1,\dots,R_N\}$ denote the segmented glomeruli and $\mathcal{P}$ the detected cells on a given section. Each glomerulus $R_k$ is labeled inflamed if its cell count $n_k = \sum_{p_i\in\mathcal{P}}\mathbf{1}[p_i\in R_k]$ exceeds threshold $\tau{=}3$, and the section-level inflamed proportion is
\begin{equation}
\rho_g = \frac{1}{N}\sum_{k=1}^{N}\mathbf{1}[n_k > \tau],
\label{eq:rho2d}
\end{equation}
The proportion $\rho_g$ is mapped to a Banff-style ordinal score: $g0$ when $\rho_g=0$, $g1$ when $0<\rho_g<25\%$, $g2$ when $25\%\leq\rho_g\leq75\%$, and $g3$ when $\rho_g>75\%$.

This module provides an automated analogue of the traditional per-section Banff assessment. Although simplified and not intended to replicate full morphological interpretation, it captures the core signal of glomerular inflammation and serves as a reproducible reference for comparison with our 3D, cross-section--aggregated scoring (Section~\ref{sec:3Dscoring}).

\subsection{Serial-Section Alignment and 3D Banff Scoring Pipeline}
\label{sec:3Dscoring}

For each biopsy case, serial histologic sections are scanned into WSIs and first processed to detect glomeruli and inflammatory cells on a per-section basis. All sections belonging to the same biopsy are aligned into a common spatial frame using a hybrid registration pipeline that combines feature-based matching and affine transformation chaining, modeling global pose variation and slice-to-slice misalignment. After alignment, glomerular masks and features from neighboring sections are compared to evaluate cross-sectional consistency, and corresponding glomerular instances exhibiting sufficient spatial overlap, centroid proximity, and morphological continuity are linked across depth. These matched instances are merged through a union--find--based matching procedure to establish stable 3D glomerular identities, which subsequently aggregate inflammatory-cell detections throughout the volume for downstream scoring.

\begin{figure*}[t]
    \centering
    \includegraphics[width=0.95\textwidth]{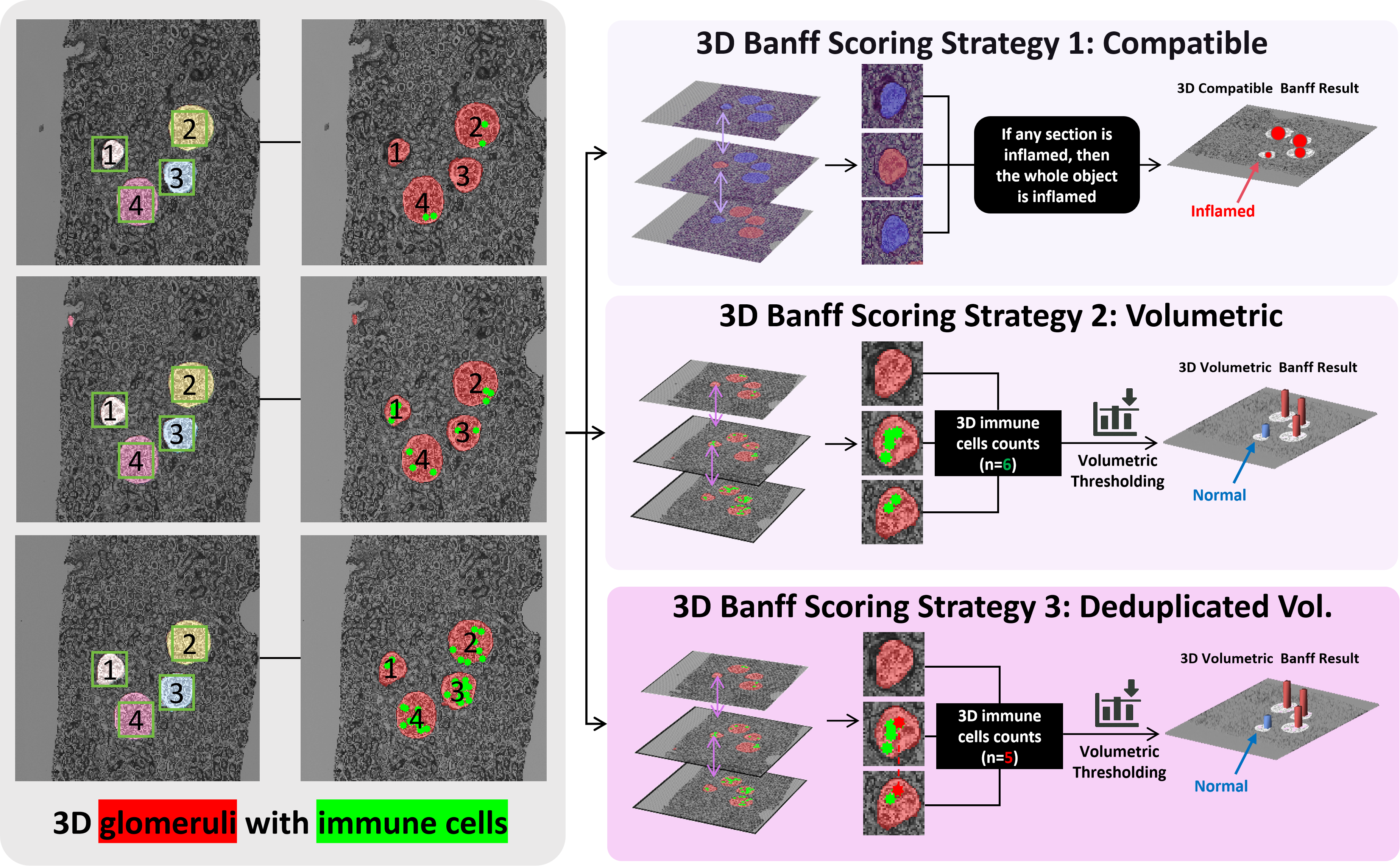}
    \caption{
    \textbf{Three 3D Banff scoring strategies enabled by cross-section glomerular alignment.}
    \textbf{Strategy~1 (Compatible):} a glomerulus is labeled inflamed if \emph{any} of its sectional instances exceeds the cell-count threshold---a direct 3D extension of the Banff worst-section rule.
    \textbf{Strategy~2 (Volumetric):} immune-cell counts are aggregated across all sections of a 3D glomerulus and compared against a section-normalized threshold, capturing the cumulative inflammatory burden.
    \textbf{Strategy~3 (Deduplicated Volumetric):} extends Strategy~2 by first deduplicating inflammatory cells that appear in multiple adjacent sections via cross-section cell-instance tracking, yielding a unique-cell volumetric count before thresholding.
    Inflamed glomeruli are shown in \textcolor{red}{red} and non-inflamed in \textcolor{blue}{blue}.
    }
    \label{fig:3d_id_assignment_pipeline}
\end{figure*}

Three distinct 3D scoring strategies are derived from these identities. For glomerulus $G_i$ spanning sections $S_i$, with raw per-section cell counts $c_{i,s}$ and cross-section-deduplicated counts $\tilde{c}_{i,s}$, each strategy defines an aggregate $C_i$ and threshold $\theta_i$:
\begin{equation}
\begin{array}{ll@{\quad}l}
\mbox{Compatible:}  & C_i = \max_{s\in S_i} c_{i,s}, & \theta_i = \tau_{\mathrm{cell}}, \\[2pt]
\mbox{Volumetric:}  & C_i = \sum_{s\in S_i} c_{i,s}, & \theta_i = \tau_{\mathrm{cell}}|S_i|, \\[2pt]
\mbox{Dedup.~Vol.:} & C_i = \textstyle\sum_{s\in S_i}\tilde{c}_{i,s}, & \theta_i = \tau_{\mathrm{cell}}|S_i|,
\end{array}
\label{eq:strategies}
\end{equation}
where $\tilde{c}_{i,s}$ removes cells already matched to the same glomerular instance in an adjacent section, and $\tau_{\mathrm{cell}}{=}3$ follows the 2D threshold. Each strategy labels $G_i$ inflamed iff $C_i \ge \theta_i$; the biopsy-level proportion $\rho_g = \#\{i : C_i \ge \theta_i\}/N$ is then mapped to ordinal Banff categories.

%% file: sections/3_experiments.tex
\section{Data and Experiments}
\subsection{Dataset}
We evaluated our framework on a retrospective cohort of renal allograft biopsies collected at Vanderbilt University Medical Center. The dataset comprises 50 patients and 410 WSIs, with each biopsy block typically sectioned into 20--30 fixed-spacing serial histological levels stained with PAS, H\&E, or Jones silver. Multiple WSIs from the same patient may correspond to different serial sections from a single biopsy block.

\subsection{Experimental Setup and Evaluation}
\label{sec:clinical_eval}
\noindent\textbf{Feature Extraction.}
Sections were detected via a YOLOv8s detector~\cite{yolov8_ultralytics} (input 1280\,px, 50 epochs, fine-tuned from COCO weights); glomeruli were segmented with the KPIs-challenge-winning pipeline~\cite{cap2025general}; and inflammatory cells were localized with a Monkey Challenge detector~\cite{Mengmonkeychallenge2023} fine-tuned on two pathologist-QA'd biopsy cases (precision 0.81, recall 0.74 on held-out regions).

\noindent\textbf{Registration and 3D Tracking.}
Serial sections were aligned using XFeat~\cite{potje2024xfeat}, following established principles for multi-gigapixel WSI registration~\cite{Gatenbee2023_VALIS}, with RANSAC-based affine estimation (threshold 4.0, max size 8192) chained relative to the middle section. Glomerular correspondences across sections were established via dilated IoU ($\geq$0.10, kernel $k{=}4$), centroid distance ($\leq$80\,px), and area ratio ($[0.5,2.0]$), merged with a Union--Find structure. Across 76 WSI cases, the pipeline produced 1,423 3D glomerular tracks (mean 2.46 serial sections per case; mean per-track section coverage: 75.2\%). All experiments used NVIDIA A6000 GPUs, Python~3.9, and PyTorch~1.12.

\noindent\textbf{Evaluation Settings.}
In the \emph{semi-automatic} setting, a board-certified pathologist performs QA on segmentation and detection outputs ($\sim$1\,min/section); all subsequent steps are fully algorithmic. In the \emph{fully automatic} setting, no human intervention is applied. Because formal expert \textit{g}-scores were not available for this retrospective cohort, clinical outcomes serve as the reference signal---a standard proxy in the absence of 3D reference standards~\cite{Loupy2022_Banff30yrs}. We report Kendall~$\tau_b$ with bootstrap 95\% CI and permutation $p$-values (10,000 draws) between predicted \textit{g}-scores and 0--6\,month changes in serum creatinine ($\Delta$sCr), blood urea nitrogen ($\Delta$BUN), and creatinine clearance ($-\Delta$Ccr) for \textbf{8} patients with longitudinal follow-up. Intra-patient WSI agreement (proportion of WSIs matching the modal \textit{g}-score per patient) is computed across the 9-patient clinical subset and visualized for all 50 patients. A sensitivity analysis was also performed by varying the glomerular matching thresholds (IoU: 0.05--0.20; centroid distance: 60--100\,px) to assess robustness of the 3D tracking pipeline.

% \begin{figure*}[t]
% \centering
% \includegraphics[width=0.9\textwidth]{Figure/QualitativeResults.png}
% \caption{
% \textbf{Qualitative comparison of glomerular inflammation scoring across three 3D Banff strategies.} \textbf{Upper panel (Case 1):} 2D worst-section scoring detects 1/25 inflamed glomeruli (4\%, g1); 3D-Compatible identifies 1/26 (3.8\%, g1); 3D-Volumetric detects 0/26 (0\%, g0). \textbf{Lower panel (Case 2):} 2D scoring finds 12/16 inflamed glomeruli (75\%, g2); 3D-Compatible and 3D-Volumetric both detect 13/16 (81.3\%, g3).
% }
% \label{fig:qualitative_combined}
% \end{figure*}

% \begin{figure}[t]
%     \centering
%     \includegraphics[width=0.72\linewidth]{Figure/MIDL_FailureCase.png}
%     \caption{
%     Representative examples of glomerulus identity tracking across serial sections.
    % \textbf{Upper Panel: Good case.} Most glomeruli are consistently matched across adjacent sections, resulting in stable identity propagation (\textcolor{green}{Green} lines).
    % \textbf{Lower Panel: Failure case.} While the majority of glomeruli are correctly associated, imperfect registration and alignment in individual sections lead to identity breaks(\textcolor{red}{Red} dashed lines and boxed regions).
%     }
%     \label{fig:failure_case}
% \end{figure}

%% file: sections/4_results.tex
\begin{figure*}[t]
    \centering
    \includegraphics[width=0.85\textwidth]{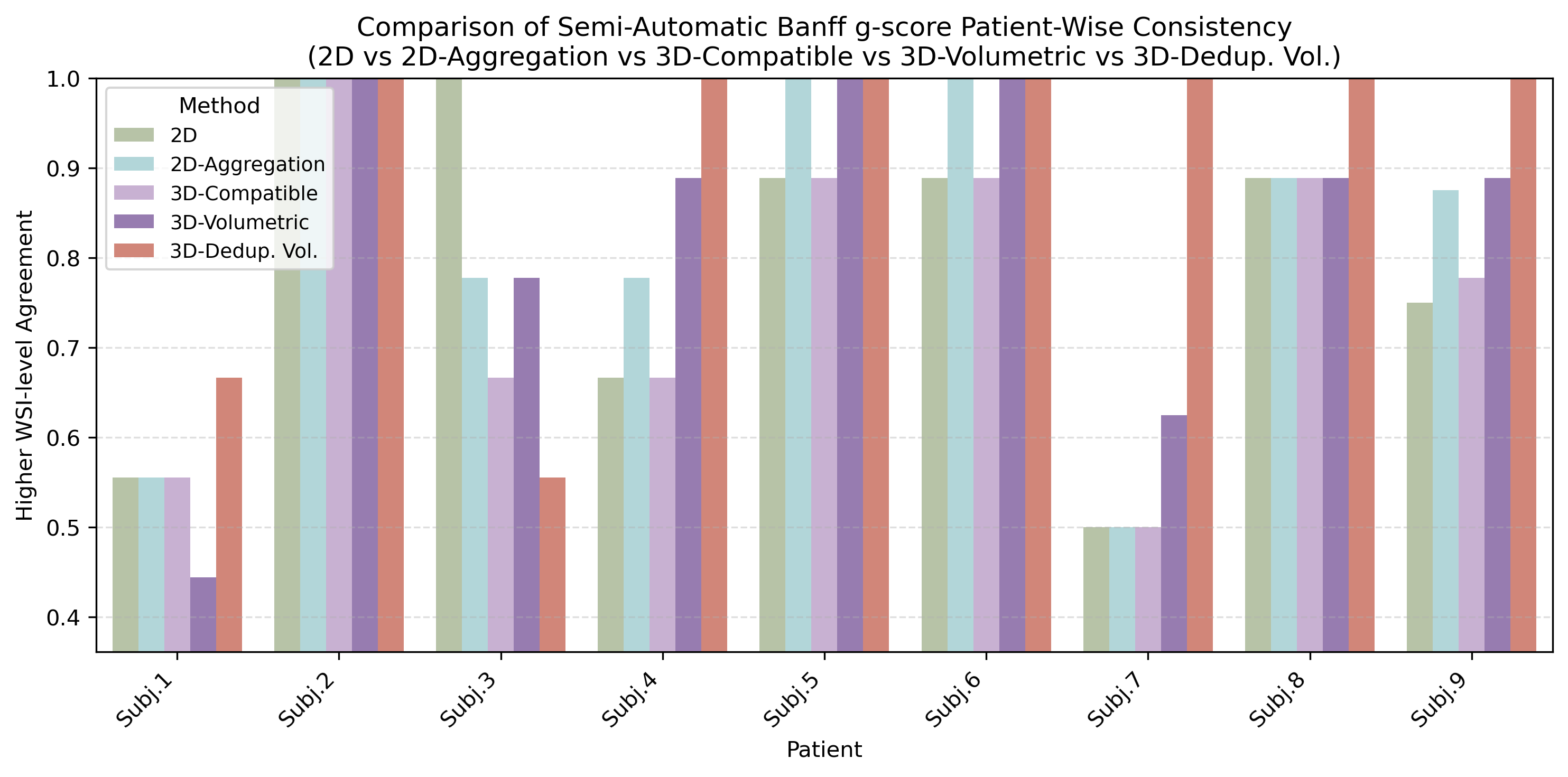}
    \caption{
    \textbf{Intra-patient WSI agreement of Banff \textit{g}-scores under Semi-Automatic pipelines.}
    For each patient, agreement is defined as the proportion of WSIs assigned the same \textit{g}-score as the modal score across that patient's sections. Five scoring strategies are evaluated: 2D worst-section scoring, 2D majority-vote aggregation (2D-Aggre.), 3D-Compatible, 3D-Volumetric, and 3D Deduplicated Volumetric (3D-Dedup.~Vol.), which further incorporates cross-section cell-instance deduplication. Higher bars represent better consistency.}
    \label{fig:patient_agreement}
\end{figure*}

\begin{figure*}[t]
\centering
\includegraphics[width=0.85\textwidth]{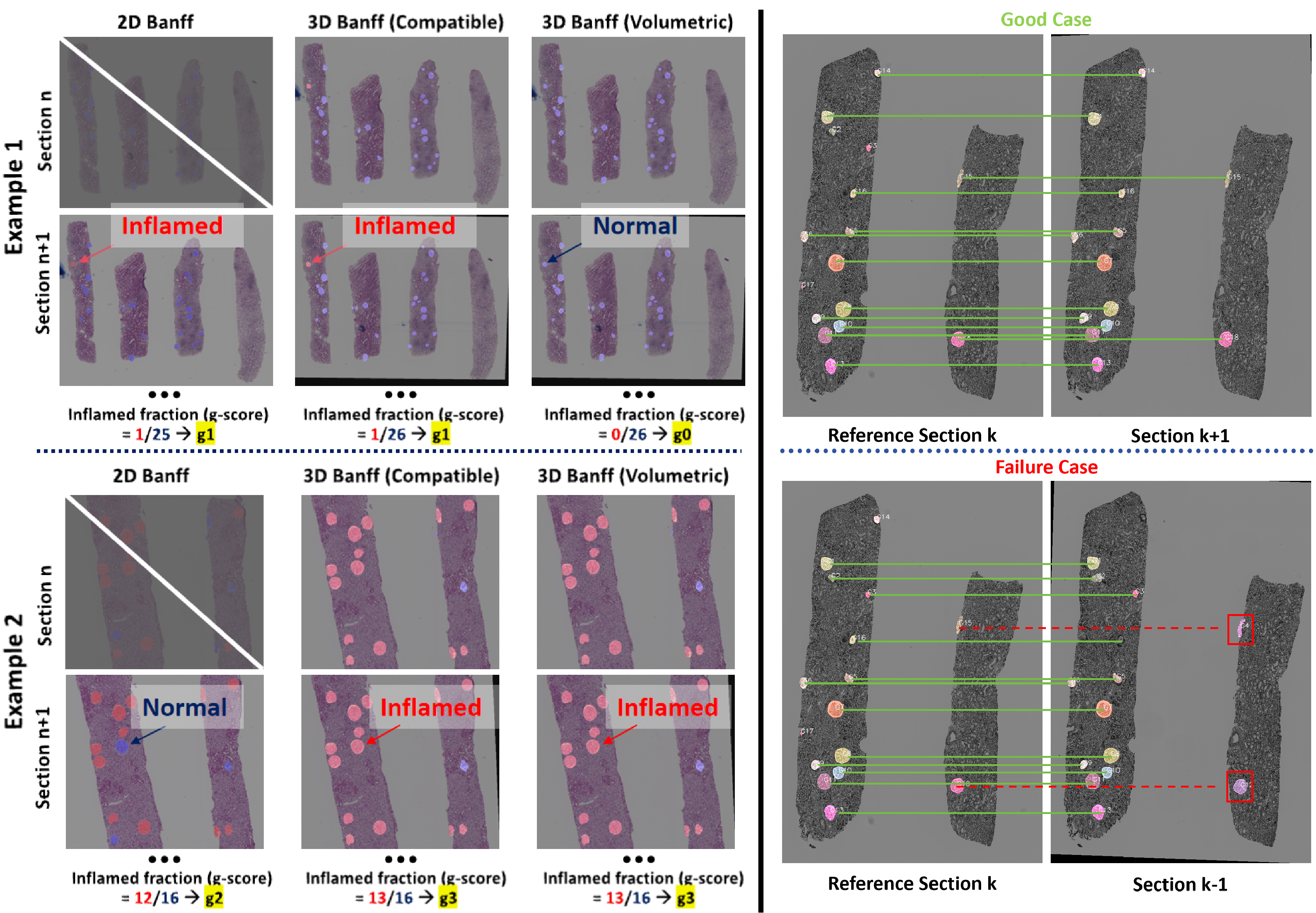}
\caption{
\textbf{Left Panel: Qualitative comparison of glomerular inflammation scoring across three 3D Banff strategies.} \textbf{Upper panel (Case 1):} 2D worst-section scoring detects 1/25 inflamed glomeruli (4\%, g1); 3D-Compatible identifies 1/26 (3.8\%, g1); 3D-Volumetric detects 0/26 (0\%, g0). \textbf{Lower panel (Case 2):} 2D scoring finds 12/16 inflamed glomeruli (75\%, g2); 3D-Compatible and 3D-Volumetric both detect 13/16 (81.3\%, g3). \textbf{Right Panel: Representative examples of glomerulus identity tracking across serial sections.}\textbf{Upper Panel: Good case.} Most glomeruli are consistently matched across adjacent sections, resulting in stable identity propagation (\textcolor{green}{Green} lines). \textbf{Lower Panel: Failure case.} While the majority of glomeruli are correctly associated, imperfect registration and alignment in individual sections lead to identity breaks(\textcolor{red}{Red} dashed lines and boxed regions).
}
\label{fig:qualitative_combined}
\end{figure*}

\begin{table}[t]
\centering
\caption{Kendall~$\tau_b$ between predicted Banff \textit{g}-scores and clinical outcomes ($n{=}8$ patients with 6-month follow-up). All differences are non-significant ($p{>}0.10$, permutation). Bold: best within each group.}
\label{tab:tau_semi_fully}
\scriptsize

% ── 2D baselines ──────────────────────────────────────────────
\begin{tabular}{l ccc ccc}
\toprule
 & \multicolumn{3}{c}{\textbf{Semi-Automatic}}
 & \multicolumn{3}{c}{\textbf{Fully-Automatic}} \\
\cmidrule(lr){2-4}\cmidrule(lr){5-7}
\textbf{2D Method} & $\Delta$sCr$\uparrow$ & $\Delta$BUN$\uparrow$ & $-\Delta$Ccr$\uparrow$
                   & $\Delta$sCr$\uparrow$ & $\Delta$BUN$\uparrow$ & $-\Delta$Ccr$\uparrow$ \\
\midrule
2D             & 0.308 & 0.077 & 0.154 & 0.227 & 0.136 & 0.045 \\
Pooled~2D\textsuperscript{*} & 0.400 & 0.109 & 0.255 & --- & --- & --- \\
Aggre.~2D      & \textbf{0.462} & \textbf{0.231} & \textbf{0.308} & 0.227 & 0.136 & 0.045 \\
\bottomrule
\end{tabular}

% ── 3D strategies ─────────────────────────────────────────────
\begin{tabular}{l ccc ccc}
\toprule
 & \multicolumn{3}{c}{\textbf{Semi-Automatic}}
 & \multicolumn{3}{c}{\textbf{Fully-Automatic}} \\
\cmidrule(lr){2-4}\cmidrule(lr){5-7}
\textbf{3D Method} & $\Delta$sCr$\uparrow$ & $\Delta$BUN$\uparrow$ & $-\Delta$Ccr$\uparrow$
                   & $\Delta$sCr$\uparrow$ & $\Delta$BUN$\uparrow$ & $-\Delta$Ccr$\uparrow$ \\
\midrule
Compat.~3D     & 0.308 & 0.076 & 0.153 & \textbf{0.233} & \textbf{0.140} & \textbf{0.047} \\
Vol.~3D        & \textbf{0.511} & \textbf{0.274} & \textbf{0.353} & \textbf{0.233} & \textbf{0.140} & \textbf{0.047} \\
Dedup.~3D      & \textbf{0.511} & \textbf{0.274} & \textbf{0.353} & \textbf{0.233} & \textbf{0.140} & \textbf{0.047} \\
\bottomrule
\end{tabular}

\vspace{1pt}
{\scriptsize \textsuperscript{*}Pooled~2D: all serial sections pooled without identity tracking; semi-auto only.}
\end{table}

\begin{figure*}[t]
    \centering
    \includegraphics[width=0.85\textwidth]{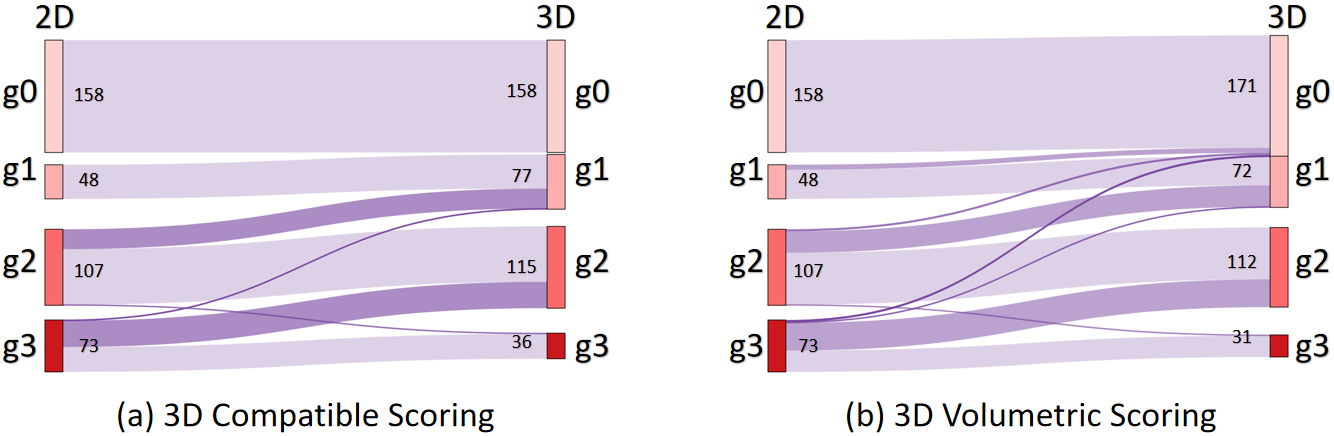}
    \caption{
        Alluvial comparison of section-level 2D \emph{g}-scores and WSI-level 3D \emph{g}-scores across 410 WSIs (50 patients) under the fully-automatic pipeline. Each WSI is assigned one 3D score; thick flows indicate frequent grade transitions. The diagram illustrates grade shifts when moving from single-section 2D scoring to:
        (a) \textbf{Compatible} scoring; and
        (b) \textbf{Volumetric} scoring.
    }
    \label{fig:banff_alluvial_results}
\end{figure*}

\section{Results}

Table~\ref{tab:tau_semi_fully} reports Kendall~$\tau_b$ correlations between predicted \textit{g}-scores and short-term clinical outcomes across five 2D/3D strategies. In the semi-automatic setting, 3D Volumetric and Dedup.~3D achieve the numerically highest correlation for $\Delta$sCr ($\tau{=}0.511$, 95\% CI $[-0.15,\,0.96]$, $p{=}0.13$). Although the small longitudinal cohort ($n{=}8$) precludes statistical significance, identity-aware 3D aggregation consistently improves over single-section 2D ($\tau{=}0.308$) and over multi-section 2D pooling without tracking (Pooled~2D: $\tau{=}0.400$), suggesting that cross-section identity tracking---not merely multi-section averaging---contributes additional discrimination. In the fully-automatic setting all 3D methods converge to $\tau{=}0.233$, with pathologist QA accounting for most of the semi-to-fully-automatic gap (mean patient-level 2D g-score disagreement: $0.39$ grade steps). Deduplication removed 1.6\% of cross-section cell detections (32/1954) and altered the inflamed status of 2/1,423 glomeruli (0.1\%), resulting in no change to any case-level ordinal g-score; both Volumetric and Dedup.~3D therefore yield identical clinical correlations in this cohort. Intra-patient WSI agreement (Figure~\ref{fig:patient_agreement}) improves from 82.9\% (2D) to 90.8\% under Dedup.~3D across the 9-patient clinical subset. Figure~\ref{fig:banff_alluvial_results} shows the \textit{g}-score distribution shift from 2D to 3D strategies across all 50 patients. Figure~\ref{fig:qualitative_combined} illustrates representative 3D scoring and tracking examples; most mismatches arise from imperfect section registration. Sensitivity analysis shows limited dependence on tracking and registration hyperparameters, with variation ratios $\leq$5.9\% and pairwise disagreement $\leq$3.9\%. The full 3D pipeline takes approximately 125\,s per WSI, a 3.34$\times$ overhead over 2D-only.

%% file: sections/5_conclusion.tex
\section{Conclusion}

We presented a feasibility study on 3D glomerulitis quantification for renal allograft biopsies, using serial-section alignment, cross-depth glomerulus identity tracking (1,423 3D tracks across 76 WSI cases), and three depth-aggregation strategies. The semi-automatic Volumetric strategy yielded the numerically strongest exploratory association with short-term renal function ($\tau{=}0.511$, 95\% CI $[-0.15,\,0.96]$ for $\Delta$sCr), outperforming both single-section 2D and multi-section 2D pooling without tracking ($\tau{=}0.400$), while Deduplicated scoring improved intra-patient WSI agreement to 90.8\%. Expert 3D reference scores and larger multi-center cohorts are needed to confirm clinical validity; the pipeline's per-WSI overhead remains practical (${\approx}125$\,s). These results establish technical feasibility and motivate prospective validation of identity-aware 3D Banff glomerulitis scoring.